\documentclass[letterpaper, 10 pt, conference]{ieeeconf}  

\IEEEoverridecommandlockouts                              

\usepackage{cite}
\usepackage{amsmath,amssymb,amsfonts}
\usepackage{algorithmic}
\usepackage{graphicx}
\usepackage{textcomp}
\usepackage{xcolor}
\usepackage{booktabs}   
\usepackage{multirow}   
\usepackage{makecell}

\title{\LARGE \bf
MCP-Net: Real-Time Coronary Perception Network and Topological Analysis for Robotic Guidewire Navigation
}

\author{Sahar Nasirihaghighi, Tianyuan Wang, Kenan Niu and Stefano Stramigioli$^{1}$
\thanks{$^{1}$ Department of Electrical Engineering, Mathematics and Computer Science (EEMCS), Robotics and Mechatronics group, University of Twente, 7500 AE Enschede, The Netherlands.
         {\tt\small s.nasirihaghighi@utwente.nl, k.niu@utwente.nl}}
}

\begin{document}

\maketitle
\thispagestyle{empty}
\pagestyle{empty}

\begin{abstract}

Coronary guidewire navigation during percutaneous coronary intervention requires continuous interpretation of vessel anatomy, guidewire position, and target stenosis under two-dimensional X-ray fluoroscopy. While existing approaches commonly address vessel segmentation or device localization as separate tasks, a robotic system requires a more structured perception of the coronary scene to support downstream navigation. In this work, we present a simulation-driven framework centered on a Multi-Task Coronary Perception Network (MCP-Net) for robotic guidewire navigation, integrating coronary vessel segmentation, vascular topology extraction, and roadmap generation. Starting from three-dimensional coronary CT angiography, we generate simulated fluoroscopy images containing vascular structures, simulated guidewires, stenotic lesions with varying severity, and diverse vascular geometries, together with simulation-derived ground truth. MCP-Net jointly performs coronary vessel segmentation, guidewire-tip localization, and stenosis localization from a single fluoroscopic frame. These predictions are subsequently converted into a structured vascular graph containing centerlines, bifurcations, endpoints, branch directions, and branch angles. By associating the predicted stenosis with this graph, the framework generates a roadmap from the automatically selected graph root toward the target lesion. The results demonstrate that the proposed framework can jointly recover coronary perception targets and transform them into an interpretable vascular representation suitable for downstream robotic navigation.

\end{abstract}

\section{Introduction}
\label{Sec:intro}
Coronary artery disease remains a major cause of morbidity and mortality worldwide, and percutaneous coronary intervention (PCI) is one of the primary procedures used to restore blood flow through stenotic coronary arteries~\cite{mensah2023global, writing20222021}. During PCI, the interventionalist must advance a thin guidewire through a complex, branching coronary vasculature toward the target lesion under two-dimensional X-ray fluoroscopic guidance. This task requires continuous interpretation of vessel geometry, identification of the lesion, and selection of the appropriate branch at successive bifurcations. The difficulty is compounded by vessel overlap, low image contrast, respiratory and cardiac motion, and the loss of depth information inherent to projection imaging~\cite{he2025deep, molenaar2025deep}. These challenges have also motivated the development of robotic-assisted PCI (RA-PCI), which enables remote manipulation of coronary guidewires and devices and has demonstrated clinical feasibility while reducing operator radiation exposure~\cite{stevenson2022robotic, haner2023robotic, bay2026robotic}. However, increasing the level of robotic assistance requires reliable intraoperative perception capable of extracting actionable vascular information directly from fluoroscopic images.

For robotic guidewire navigation, however, vessel segmentation alone is not sufficient. A navigation system must reason about both the device and its surrounding anatomy: where the guidewire tip is located, where the target stenosis lies, how the coronary tree branches, and which sequence of vessel segments connects the coronary root to the target. Such information must also be available at interactive rates if it is to support intraoperative robotic decision-making. Existing work has investigated coronary vessel segmentation, guidewire or catheter localization, and vascular analysis~\cite{nobre2023coronary, zhou2020pyramid, ma2020dynamic}, but these components are commonly considered as separate problems. Consequently, even an accurate segmentation mask does not directly provide the structured anatomical representation required for downstream navigation, including vessel connectivity, bifurcation structure, branch orientation, and a path toward the target lesion.

A second challenge is the availability of suitable training and evaluation data. Dense annotation of coronary fluoroscopy is difficult and time-consuming, particularly for thin or overlapping vessels. Precise annotation of guidewire tips, stenosis location, and stenosis severity introduces additional ambiguity, while annotations of centerlines and vascular topology require further expert effort. More importantly, the ground-truth geometry underlying a two-dimensional projection is generally unavailable in real fluoroscopic data. These limitations make it difficult to jointly train and quantitatively evaluate all of the perception and geometric components required by a robotic navigation pipeline.

In this work, we address these challenges through a simulation-driven, multi-task framework for coronary perception, topological analysis, and real-time roadmap generation. Starting from three-dimensional coronary CT angiography, we generate fluoroscopy-like projections containing coronary vessels, simulated guidewires, and stenotic lesions with different severity levels across diverse vascular geometries. Since the underlying scene geometry is known, task-specific ground truth can be generated automatically for vessel segmentation, guidewire-tip localization, stenosis localization, and geometric analysis. These data are used to train the proposed Multi-Task Coronary Perception Network (MCP-Net), which jointly predicts the vessel mask, guidewire-tip location, and stenosis landmarks from a single fluoroscopy frame. The predicted vasculature is subsequently converted into a structured centerline graph containing bifurcations, endpoints, vessel geometry, and branch information, from which a target-oriented vascular roadmap is generated.

This design deliberately separates learned image perception from geometric vascular reasoning. MCP-Net performs the image-dependent tasks jointly in a single forward pass, while centerline extraction, graph construction, and roadmap generation operate directly on the resulting vascular representation. The combination provides not only pixel-level predictions, but also a compact and interpretable description of the coronary anatomy that can serve as a perception interface for downstream robotic planning.

The main contributions of this work are:
\begin{itemize}

    \item We develop a CT-based coronary fluoroscopy simulation framework that generates synthetic intervention scenes containing coronary vasculature, simulated guidewires, and stenotic lesions with varying severity, together with automatically generated task-specific ground truth.

    \item We propose MCP-Net, a multi-task coronary perception network that jointly performs coronary vessel segmentation, guidewire-tip localization, and stenosis localization within a unified architecture.

    \item We introduce a real-time vascular topology and roadmap generation framework that converts the predicted vasculature into a structured graph containing centerlines, vessel calibers, bifurcations, endpoints, and branch geometry, and generates a navigation-oriented path toward the detected stenosis.

    \item We provide a comprehensive evaluation on the simulated fluoroscopy dataset, including comparisons with established segmentation architectures, vessel segmentation and connectivity, guidewire-tip and stenosis localization, lesion geometry, vascular topology, roadmap generation, and end-to-end computational performance.

\end{itemize}

The remainder of this paper is organized as follows. Section~\ref{sec:relatedwork} reviews related work, and Section~\ref{sec:methods} presents the proposed framework. Sections~\ref{sec:setups} and~\ref{sec:results} describe the experimental setup and results, respectively. Section~\ref{sec:discussion} discusses the findings, limitations, and future directions, and Section~\ref{sec:conclusion} concludes the paper.

\section{Related Work}
\label{sec:relatedwork}
\subsection{Coronary Vessel Segmentation in X-ray Angiography}

Deep learning has substantially improved coronary vessel segmentation in X-ray angiography, with recent work focusing on vessel continuity, temporal information, and reducing dependence on manual annotations. He et al. proposed TVS-Net, which incorporates temporal information from angiographic sequences together with a connectivity-preserving objective~\cite{he2025deep}. Li et al. further introduced a subpixel-level framework for X-ray coronary angiography, combining vessel landmark localization with topology-aware contour estimation for quantitative vessel analysis~\cite{li2025straightening}.

The difficulty of obtaining dense annotations from clinical angiography has also motivated simulation- and self-supervision-based approaches. Zhang et al. used digitally reconstructed radiographs generated from CT data together with domain randomization and self-supervised learning to train coronary vessel segmentation without manual angiographic annotations~\cite{zhang2024self}. More recently, XA-Sim2Real generates simulated X-ray angiograms from coronary CT angiography and learns adaptive representations between simulated and real images~\cite{zhang2024xa}. These approaches demonstrate the value of CT-derived supervision, but primarily target vessel segmentation. In contrast, our simulation pipeline additionally models the interventional guidewire and stenotic lesion, providing simulation-derived supervision for multiple perception tasks and subsequent vascular topology analysis.

\subsection{Guidewire and Lesion Localization}

Reliable localization of interventional devices is another important component of image-guided vascular procedures. Zhang et al. proposed a two-stage real-time framework for guidewire tracking and segmentation in intraoperative X-ray images using spatial-temporal refinement and guidewire-specific feature enhancement \cite{zhang2022real}. Yao et al. subsequently introduced a Siamese attention network for real-time guidewire-tip tracking~\cite{yao2025real}. Recent work by Robertshaw et al. integrated catheter/guidewire segmentation, skeletonization, and tip extraction into a real-time fluoroscopic tracking pipeline intended for downstream autonomous navigation~\cite{robertshaw2026toward}. Multi-task learning has also been explored for interventional device perception; Xi et al. jointly performed catheter segmentation and electrode localization using a shared network with multiple prediction heads~\cite{xi2026catheter}.

Stenosis detection has largely been studied separately from device perception. DeepCoro employs vessel tracking and video-based models for automatic localization and assessment of coronary stenoses~\cite{labrecque2024evaluation}, while Molenaar et al. recently developed and externally validated deep-learning models for both coronary artery segmentation and significant stenosis detection in invasive coronary angiography~\cite{molenaar2025deep}. In contrast to these task-specific or paired formulations, MCP-Net jointly estimates the coronary vessel mask, guidewire-tip location, and stenosis location from a single fluoroscopy frame using a shared representation.

\subsection{Vascular Topology and Robotic Endovascular Navigation} 
Beyond pixel-level segmentation, structured vascular representations are important for anatomical reasoning and robotic navigation. Hampe et al. represented coronary artery trees as graphs for automatic extraction and anatomical labeling in coronary CT angiography~\cite{hampe2024graph}. Such centerline- and graph-based representations provide explicit information about vessel connectivity and branching structure that is not directly available from a binary segmentation mask.

\begin{figure*}[!t]
    \centering
    \includegraphics[width=0.9\textwidth]{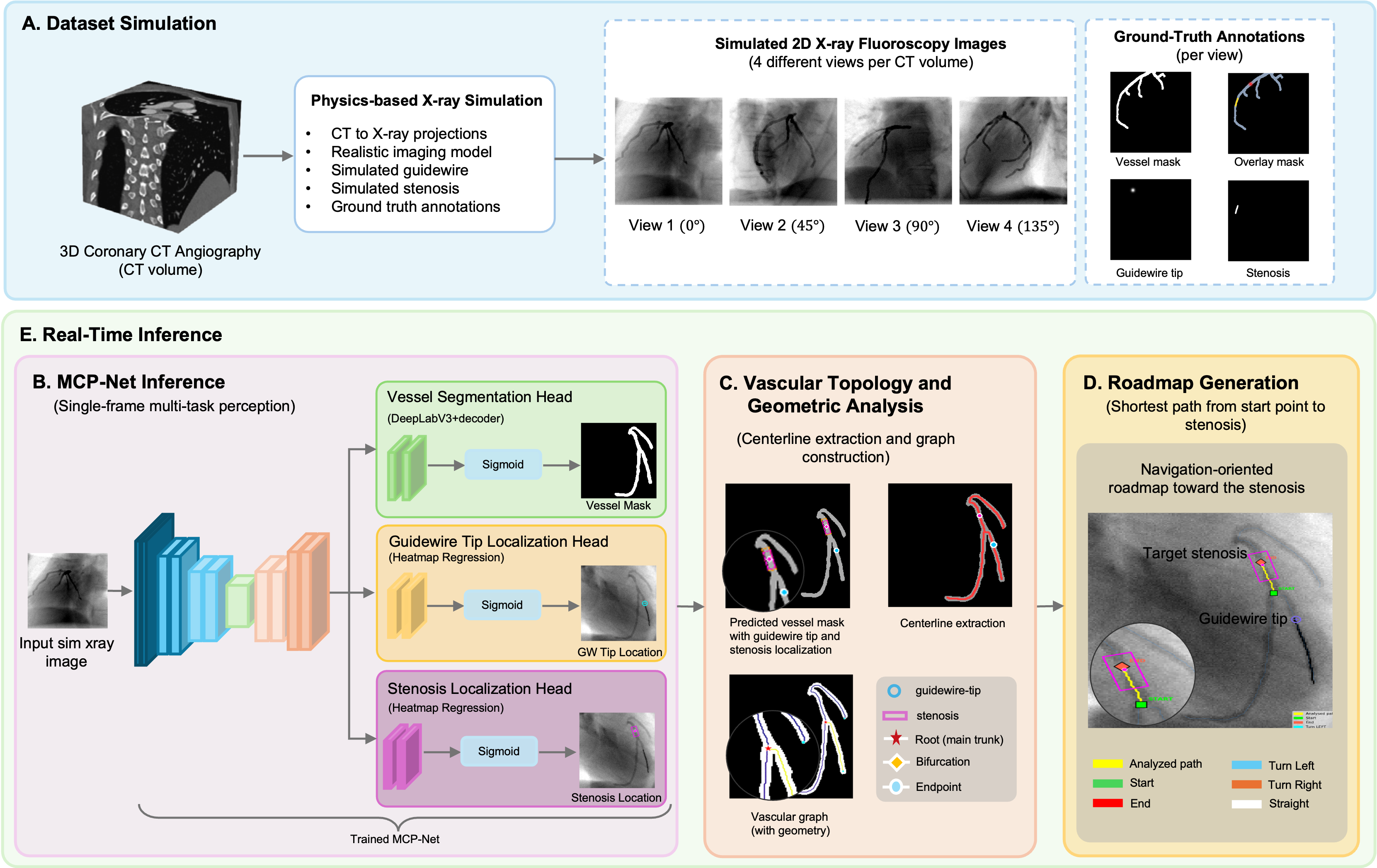}
    \caption{Overview of the proposed framework. Coronary CT angiography is used to generate simulated fluoroscopy images and task-specific annotations for MCP-Net training. During real-time inference, the trained MCP-Net jointly predicts the vessel mask, guidewire-tip location, and stenosis landmarks from a single fluoroscopy frame. The predicted vasculature is converted into a structured centerline graph and combined with the predicted stenosis target to generate a navigation-oriented roadmap.}
    \label{fig:framework}
\end{figure*}

In robotic endovascular intervention, Li et al. proposed a vascular centerline-guided autonomous navigation method, demonstrating the utility of centerline geometry for robot-assisted navigation~\cite{li2024vascular}. Scarponi et al. further investigated reinforcement learning for autonomous guidewire navigation in unseen vascular geometries~\cite{scarponi2024zero} and extended this direction to deforming anatomy under single-view fluoroscopic imaging~\cite{scarponi2024autonomous}. These works highlight the importance of reliable geometric and device state information for robotic navigation. Our work focuses on producing such information directly from fluoroscopic perception: the predicted vessel mask is converted into a structured graph containing centerlines, bifurcations, endpoints, vessel calibers, and branch geometry, from which a root-to-lesion roadmap is derived. 

\section{Methods}
\label{sec:methods}

The proposed framework consists of an offline simulation and training stage followed by a real-time perception and vascular-analysis pipeline, as illustrated in Fig.~\ref{fig:framework}. First, simulated fluoroscopy images and their task-specific annotations are generated from coronary CT angiography (Section~III-A) and used to train MCP-Net. During real-time inference, the trained MCP-Net processes each incoming fluoroscopy frame in a single forward pass to predict the coronary vessel mask, guidewire-tip location, and stenosis landmarks (Section~III-B). The predicted vessel mask is subsequently converted into a centerline-based vascular graph containing topological and geometric information (Section~III-C). The graph is then combined with the localized target to generate a navigation-oriented roadmap toward the stenosis (Section~III-D). Section~III-E summarizes the complete
real-time inference pipeline. 

\subsection{Data Simulation}

Real fluoroscopy with dense, per-pixel and per-point annotations is scarce and impractical to label by hand. We therefore synthesize training data from three-dimensional coronary CT angiography~\cite{zeng2023imagecas}. For each CT volume, multiple two-dimensional fluoroscopy-like projections are generated using Tomosipo~\cite{hendriksen2021tomosipo}, with the underlying projection operations performed by the ASTRA Toolbox~\cite{van2015astra}. Vessel centerlines and local radius profiles are extracted using the Vascular Modeling Toolkit (VMTK)~\cite{izzo2018vascular}. Stenotic lesions are simulated by locally reducing the vessel radius along a selected centerline segment, allowing lesions with different severity levels and locations to be generated. A simulated guidewire is subsequently introduced and advanced toward the stenotic region using the coronary centerline as a geometric reference. Because the underlying anatomy and simulated intervention components are known, exact simulation-derived annotations can be generated automatically for each projection. These include the coronary vessel mask, guidewire-tip location, and stenosis landmarks, together with additional geometric information required for downstream analysis. This simulation stage is performed offline and provides the dataset used to train MCP-Net and evaluate the complete framework.

\subsection{Multi-Task Coronary Perception Network (MCP-Net)}
Given a single fluoroscopy frame $I$, MCP-Net jointly predicts the coronary vessel mask, guidewire-tip location, and stenosis landmarks. A shared feature representation is followed by three task-specific prediction heads for vessel segmentation, guidewire-tip localization, and stenosis localization. The vessel branch employs a DeepLabV3+ decoder with atrous spatial pyramid pooling (ASPP) to produce a dense binary coronary segmentation. The vessel branch is optimized using the combination of binary cross-entropy and Dice losses,

\begin{equation}
    \mathcal{L}_{\mathrm{seg}}
    = \mathcal{L}_{\mathrm{BCE}} + \mathcal{L}_{\mathrm{Dice}},
\end{equation}

which balances pixel-wise classification with region overlap and is well suited to the sparse, thin vessel structures.

The two localization branches predict spatial heatmaps for the guidewire tip and stenosis landmarks. Rather than treating localization as pixel-wise classification, we recover continuous image coordinates from each predicted heatmap using a differentiable spatial soft-argmax (DSNT),

\begin{equation}
    \hat{\mathbf{u}}
    = \sum_{\mathbf{p}} \mathbf{p}\,
      \operatorname{softmax}(\hat{H})_{\mathbf{p}} .
\end{equation}

where $\hat{H}$ denotes the predicted heatmap and $\mathbf{p}$ denotes a pixel coordinate. The guidewire branch predicts a single tip location, whereas the stenosis branch predicts three landmarks corresponding to the proximal boundary, lesion throat, and distal boundary, thereby representing both the location and extent of the stenosis. Localization losses are applied only to valid annotated targets.

The complete network is trained end-to-end with the joint objective
\begin{equation}
    \mathcal{L}
    = \mathcal{L}_{\mathrm{seg}}
    + \lambda_{\mathrm{tip}}\mathcal{L}_{\mathrm{tip}}
    + \lambda_{\mathrm{sten}}\mathcal{L}_{\mathrm{sten}},
\end{equation}
where $\lambda_{\mathrm{tip}}=\lambda_{\mathrm{sten}}=5$ in our experiments. At run time, only the trained MCP-Net checkpoint is used, and all three perception outputs are obtained in a single forward pass.

\subsection{Vascular Topology and Geometric Analysis}

The predicted vessel mask is converted into a structured vascular representation using deterministic geometric processing. Small disconnected components are first removed, after which morphological skeletonization reduces the vessel tree to a one-pixel-wide centerline. A Euclidean distance transform of the binary vessel mask is used to estimate the local vessel caliber. For a centerline point $\mathbf{p}$, the local diameter is defined as

\begin{equation}
    d(\mathbf{p}) = 2D(\mathbf{p}),
\end{equation}

where $D(\mathbf{p})$ is the Euclidean distance from $\mathbf{p}$ to the
nearest background pixel.

The extracted centerline is subsequently converted into a vascular graph $\mathcal{G}=(\mathcal{V},\mathcal{E})$. Graph nodes represent vessel bifurcations and terminal endpoints, while edges correspond to the centerline segments connecting them. Each edge retains its centerline trajectory together with geometric attributes including segment length and local vessel diameter. Short spurious segments are removed and nearby junctions are consolidated to reduce artifacts introduced by discrete skeletonization.

Local vessel directions are estimated from the centerline geometry around each junction and are used to compute bifurcation and take-off angles. The resulting graph therefore captures both the connectivity and local geometry of the predicted coronary tree and provides the structured representation required for subsequent landmark association and roadmap generation.

\subsection{Roadmap Generation}
The vascular graph and predicted stenosis location are combined to construct an automatic navigation roadmap toward the target lesion. The start point is defined by the automatically selected graph root, which serves as a proxy for the coronary ostium, while the predicted stenosis throat defines the target.

Both locations are projected onto the nearest vessel centerline points. If either lies within a graph edge, a temporary node is inserted at the projected location to preserve the exact start and end positions. A shortest-path search is then performed using vessel-segment arc length as the edge weight, producing a continuous centerline path from the root to the stenosis.

Along the resulting path, traversed bifurcations are identified and local vessel directions are used to compute branch angles and classify each navigation decision as left, right, or straight. The resulting roadmap therefore provides both the target-oriented vessel path and bifurcation-level navigation cues for downstream robotic planning or guidewire control.

\subsection{Real-Time Inference}
During online operation, each fluoroscopy frame is processed by the trained MCP-Net in a single forward pass to obtain the vessel mask, guidewire-tip location, and stenosis location.

The predicted vessel mask is then converted into a centerline-based vascular graph containing the root, bifurcations, endpoints, vessel calibers, and branch geometry. The predicted stenosis throat is associated with this graph, and the roadmap-generation procedure in Section~III-D is applied automatically.

The final output consists of the root-to-stenosis path, path length, traversed bifurcations, branch angles, and left/right/straight navigation cues. These outputs can be overlaid with the guidewire-tip and stenosis predictions on the fluoroscopy image. No network training or manual landmark selection is performed during online inference.

\section{Experimental Setups}
\label{sec:setups}

\subsection{Dataset}
The simulated fluoroscopy dataset is generated from coronary CT angiography volumes of ImageCAS~\cite{zeng2023imagecas} using the procedure described in Section~III-A. In total, 376 CT volumes are used, with four simulated projection views generated per volume, resulting in 1,504 images of size $256\times256$ pixels. The split is performed at the CT-volume level to avoid anatomical overlap between sets: 208 volumes (832 images) are used for training, 84 volumes (336 images) for validation, and 84 volumes (336 images) for testing.

Each image is accompanied by simulation-derived annotations including a binary vessel mask, guidewire-tip heatmap, and three stenosis landmarks corresponding to the proximal boundary, throat, and distal boundary. These annotations provide supervision for all perception tasks without manual fluoroscopy labeling.

\subsection{Implementation Details}
MCP-Net uses DeepLabV3+ with a ResNet-50 backbone and $256\times256$ inputs. Training is performed for 40 epochs with batch size 4 using SGD ($\mathrm{lr}=5\times10^{-3}$, momentum $0.9$, weight decay $10^{-4}$) and cosine annealing to $10^{-5}$. The segmentation branch uses BCE+Dice loss, while the guidewire-tip and stenosis branches use DSNT-based localization losses with $\lambda_{\mathrm{tip}}=\lambda_{\mathrm{sten}}=5$.

At inference, vessel probabilities are thresholded at 0.5, while landmark coordinates are recovered using an $11\times11$ local soft-argmax window. The predicted vessel mask is subsequently skeletonized and converted into a vascular graph, from which vessel connectivity, local caliber, and branch geometry are extracted. For the architecture comparison, DeepLabV3+~\cite{chen2018encoder}, U-Net~\cite{ronneberger2015u}, SegFormer~\cite{xie2021segformer}, and FPN~\cite{lin2017feature} use identical multi-task heads, losses, data splits, and training schedules. All experiments were performed on an NVIDIA GB10 GPU with CUDA 13.0.

\subsection{Evaluation Metrics}
Vessel segmentation is evaluated using Dice, IoU, and centerline Dice (clDice). Guidewire-tip and stenosis-throat localization are evaluated using Euclidean pixel error, reported as median and interquartile range (IQR), together with PCK@5 and PCK@10. For stenosis localization, we additionally report the fraction of predicted throat points lying within the ground-truth lesion.

Lesion extent is evaluated using the absolute and relative error of the proximal-to-distal landmark distance. For the subset of stenosis predictions with throat error $\leq 8$ pixels, we additionally report the median localization error. Computational efficiency is evaluated using the end-to-end latency and speed of the complete perception, topology, and roadmap pipeline.
\section{Experimental Results}
\label{sec:results}

Table~\ref{tab:deeplabv3_results} summarizes the test-set performance of the final MCP-Net configuration using DeepLabV3+. For vessel segmentation, it achieved a Dice score of $90.3\pm3.0\%$, an IoU of $82.5\pm4.8\%$, and a clDice score of $96.0\pm3.0\%$. The higher clDice relative to the region-overlap metrics indicates that the predicted masks
preserve the centerline structure of the coronary tree particularly well. This property is important for the subsequent skeletonization and graph construction, where discontinuities in thin vessel branches can directly affect the extracted vascular topology.

\begin{table}[t]
\centering
\caption{Performance of the DeepLabV3+-based MCP-Net on the simulated test
set ($n=336$). Localization errors are reported as median (IQR).}
\label{tab:deeplabv3_results}
\footnotesize
\setlength{\tabcolsep}{3.5pt}
\begin{tabular}{@{}llr@{}}
\toprule
Task & Metric & Value \\
\midrule
\multirow{3}{*}{Vessel segmentation}
& Dice   & $90.3 \pm 3.0\%$ \\
& clDice & $96.0 \pm 3.0\%$ \\
& IoU    & $82.5 \pm 4.8\%$ \\
\midrule
\multirow{2}{*}{Guidewire tip ($n=327$)}
 & Error & $1.09$ ($0.73$--$1.95$) px \\
 & PCK@5 & $88.1\%$ \\
\midrule
\multirow{4}{*}{Stenosis throat ($n=320$)}
 & Error & $3.82$ ($1.88$--$27.1$) px \\
 & PCK@5 / PCK@10 & $52.7 / 64.4\%$ \\
 & Detection rate & $70.6\%$ \\
 & Error if $\leq8$ px & $2.13$ ($1.29$--$3.23$) px \\
\midrule
\multirow{2}{*}{Lesion length}
 & Absolute error & $4.60$ ($1.48$--$11.30$) px \\
 & Relative error & $17.7\%$ \\
\bottomrule
\end{tabular}
\end{table}

\begin{figure}[t]
    \centering
    \includegraphics[width=0.38\textwidth]{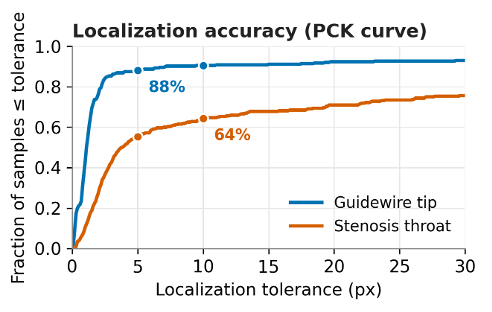}
    \caption{Localization accuracy as a function of pixel tolerance for guidewire-tip and stenosis-throat localization on the simulated test set. }
    \label{fig:pck}
\end{figure}

Guidewire-tip localization was accurate for most valid test samples, achieving a median error of $1.09$ px (IQR: $0.73$--$1.95$ px) and a PCK@5 of $88.1\%$. The narrow IQR indicates that the majority of tip predictions are concentrated close to the ground-truth location. In comparison, stenosis-throat localization was more challenging, with a median error of $3.82$ px (IQR: $1.88$--$27.1$ px), PCK@5 of $52.7\%$, and PCK@10 of $64.4\%$. The considerably larger upper quartile indicates the presence of a subset of localization errors.

Despite this wider error distribution, $70.6\%$ of predicted throat locations fell within the ground-truth lesion extent. This suggests that some predictions identify the correct stenotic region without precisely localizing the lesion throat. For the subset of predictions with throat error $\leq8$ px, the median error decreased to $2.13$ px (IQR: $1.29$--$3.23$ px), showing substantially tighter localization among well-localized cases. The proximal and distal stenosis landmarks additionally provided an estimate of lesion extent. Across the valid lesion-length cases ($n=316$), the proximal-to-distal distance had a median absolute error of $4.60$ px and a median relative error of $17.7\%$.

\begin{table*}[t]
\centering
\caption{Comparison of multi-task architectures on the simulated test set ($n=336$). Best results are shown in bold.}
\label{tab:arch}
\small
\setlength{\tabcolsep}{4.5pt}
\begin{tabular}{@{}lccccccccc@{}}
\toprule
Network &
Dice (\%) &
clDice (\%) &
IoU (\%) &
\makecell{Tip error\\(px) $\downarrow$} &
\makecell{Tip\\PCK@5 $\uparrow$} &
\makecell{Throat error\\(px) $\downarrow$} &
\makecell{Throat\\PCK@10 $\uparrow$} &
\makecell{Throat in\\lesion (\%) $\uparrow$} &
\makecell{Error $>$15 px\\(\%) $\downarrow$} \\
\midrule

DeepLabV3+
& \textbf{90.3}
& \textbf{96.0}
& \textbf{82.5}
& \textbf{1.09}
& \textbf{88.1}
& \textbf{3.82}
& \textbf{64.4}
& 70.6
& \textbf{32.2} \\

U-Net
& 87.7
& 94.4
& 78.3
& 1.79
& 82.3
& 5.92
& 62.2
& \textbf{71.5}
& 34.7 \\

SegFormer
& 86.5
& 93.1
& 76.5
& 2.06
& 74.3
& 6.37
& 57.5
& 68.0
& 35.0 \\

FPN
& 85.7
& 93.4
& 75.2
& 1.54
& 79.5
& 8.97
& 53.1
& 66.5
& 41.6 \\

\bottomrule
\end{tabular}
\end{table*}

\begin{figure*}[!t]
    \centering
    \includegraphics[width=1\textwidth]{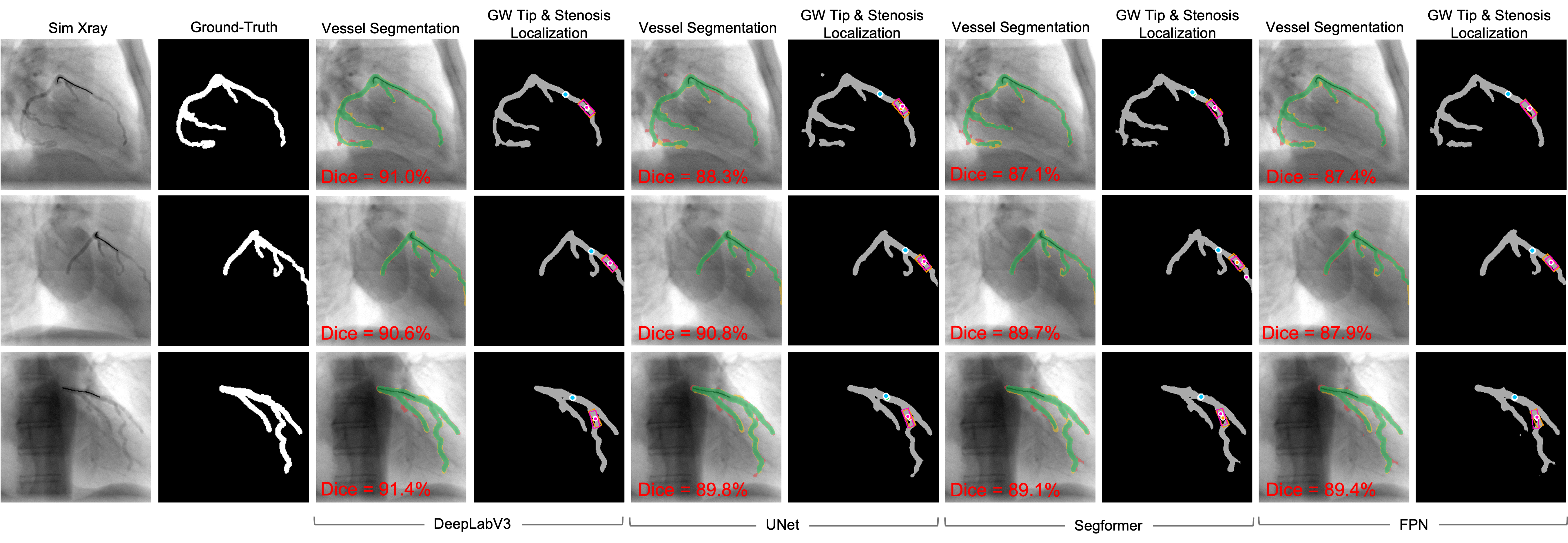}
    \caption{Qualitative comparison of the multi-task architectures on representative simulated fluoroscopy images. For each test case, the input image and ground-truth vessel mask are followed by vessel-segmentation and guidewire-tip/stenosis localization predictions from DeepLabV3+, U-Net, SegFormer, and FPN.}
    \label{fig:network_comparison}
\end{figure*}

Fig.~\ref{fig:pck} further illustrates the different localization behavior of the two tasks. Guidewire-tip accuracy rises rapidly at small pixel tolerances and then approaches a plateau, consistent with its relatively compact error distribution. In contrast, stenosis-throat localization improves more gradually as the tolerance increases, reflecting the broader error distribution observed in Table~\ref{tab:deeplabv3_results}. Together, these results indicate that guidewire-tip localization is the more stable of the two localization tasks, while precise identification of the stenosis throat remains more challenging.

Table~\ref{tab:arch} compares DeepLabV3+, U-Net, SegFormer, and FPN using the same multi-task formulation and training conditions. DeepLabV3+ achieved the highest vessel-segmentation performance, with Dice, clDice, and IoU scores of $90.3\%$, $96.0\%$, and $82.5\%$, respectively. It also obtained the lowest median guidewire-tip error ($1.09$ px) and stenosis-throat error ($3.82$ px), together with the lowest proportion of large throat-localization errors ($>15$ px).

The alternative architectures showed different trade-offs. FPN produced the second-lowest guidewire-tip error ($1.54$ px), but showed the largest stenosis-throat error ($8.97$ px) and the highest proportion of errors above 15 px. U-Net was comparatively competitive for stenosis localization and achieved a slightly higher fraction of predictions within the ground-truth lesion than DeepLabV3+ ($71.5\%$ versus $70.6\%$), although its median throat error remained higher ($5.92$ px). Considering both vessel segmentation and landmark localization performance, DeepLabV3+ was selected for the final MCP-Net configuration.

The qualitative comparison in Fig.~\ref{fig:network_comparison} is consistent with the quantitative results. The same fluoroscopy cases are processed by all four architectures, enabling direct comparison of vessel segmentation and landmark localization. Across the examples, DeepLabV3+ maintains the major vessel structure and thin branches more consistently, while also preserving the guidewire-tip and stenosis predictions required by the subsequent analysis. The comparison also illustrates that differences between architectures are not limited to pixel-wise vessel overlap, but can influence the vascular structure available for downstream graph extraction.

\begin{figure}[!t]
    \centering
    \includegraphics[width=0.45\textwidth]{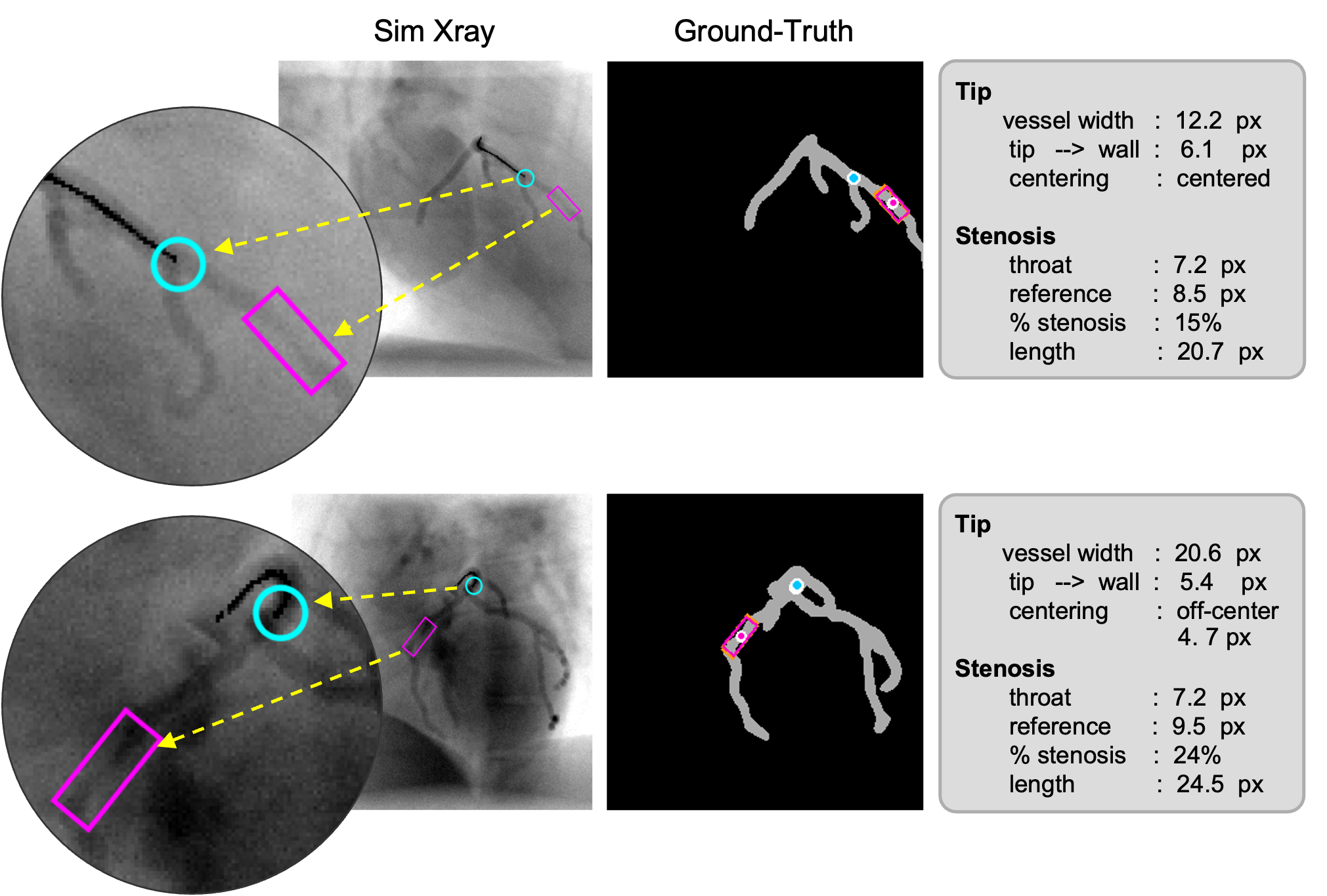}
    \caption{Representative guidewire-tip and stenosis geometric analysis. Predicted landmarks are associated with the vessel geometry to estimate guidewire-tip position relative to the vessel, local vessel width, stenosis throat and reference diameters, lesion extent, and descriptive image-based percentage diameter stenosis (\%DS).}
    \label{fig:gw_stenosis_loc}
\end{figure}

Beyond point localization, the MCP-Net outputs are associated with the predicted vessel geometry to obtain additional navigation-relevant information. Representative examples are shown in Fig.~\ref{fig:gw_stenosis_loc}. For the guidewire tip, the framework relates the predicted location to the vessel centerline and estimates the local vessel width. This analysis indicates whether the guidewire tip remains near the vessel centerline or deviates toward the vessel boundary, providing useful feedback for downstream guidewire navigation. For the stenosis, the proximal, throat, and distal landmarks provide estimates of lesion extent, throat diameter, and local reference diameter, and percentage diameter stenosis.

\begin{figure}[t]
    \centering
    \includegraphics[width=0.48\textwidth]{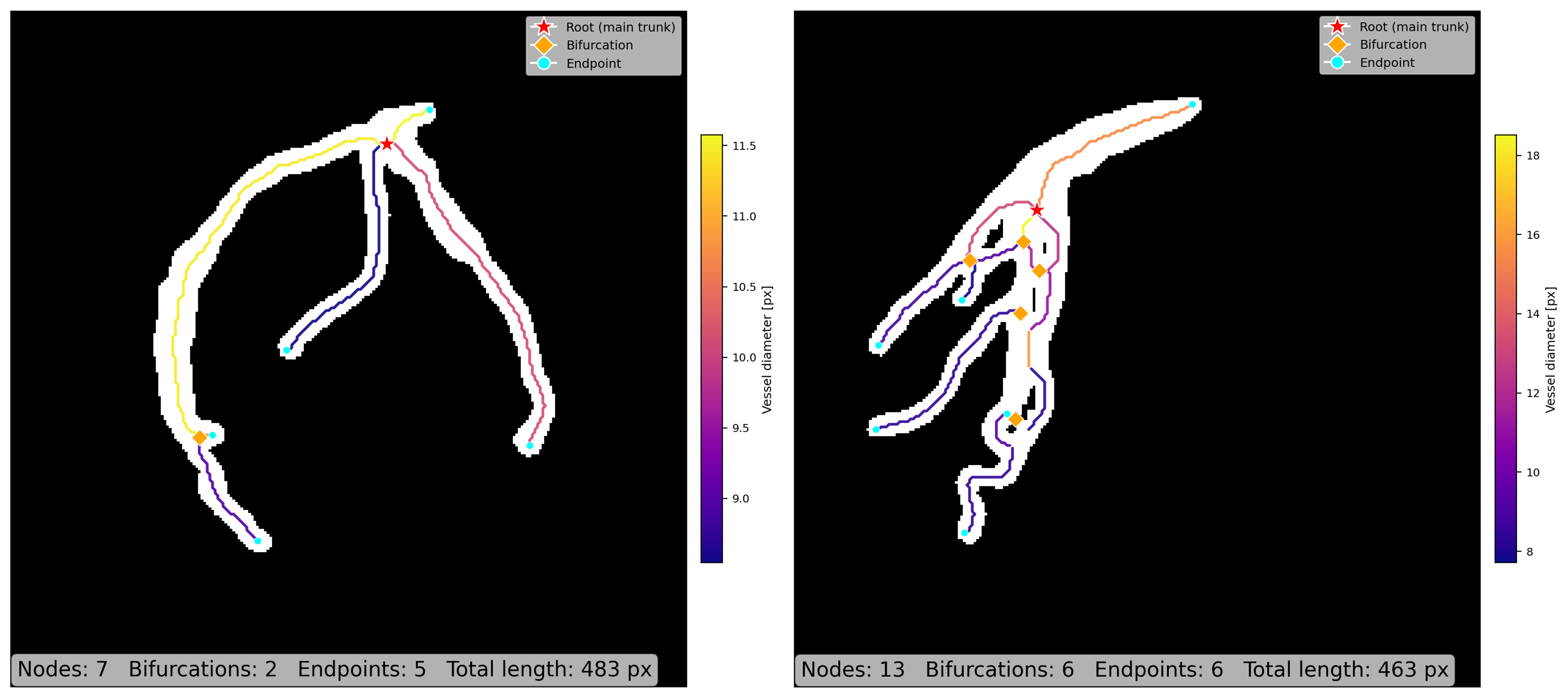}
    \caption{Representative vascular topology extracted from MCP-Net vessel predictions. The predicted mask is skeletonized and converted into a graph containing the automatically selected root, bifurcations, endpoints, and connecting vessel segments. Graph edges are color-coded according to their vessel diameter.}
    \label{fig:topology}
\end{figure}

Fig.~\ref{fig:topology} qualitatively demonstrates the vascular topology obtained from the predicted vessel masks. The centerline representation is converted into a graph containing the selected root, bifurcations, endpoints, and connecting vessel segments, while the graph edges retain local geometric information such as vessel caliber. The two examples show that the same processing pipeline can represent coronary trees with different branching complexity. This structured representation provides the connectivity and branch geometry required for the subsequent root-to-stenosis path analysis.

\begin{figure}[t]
    \centering
    \includegraphics[width=0.45\textwidth]{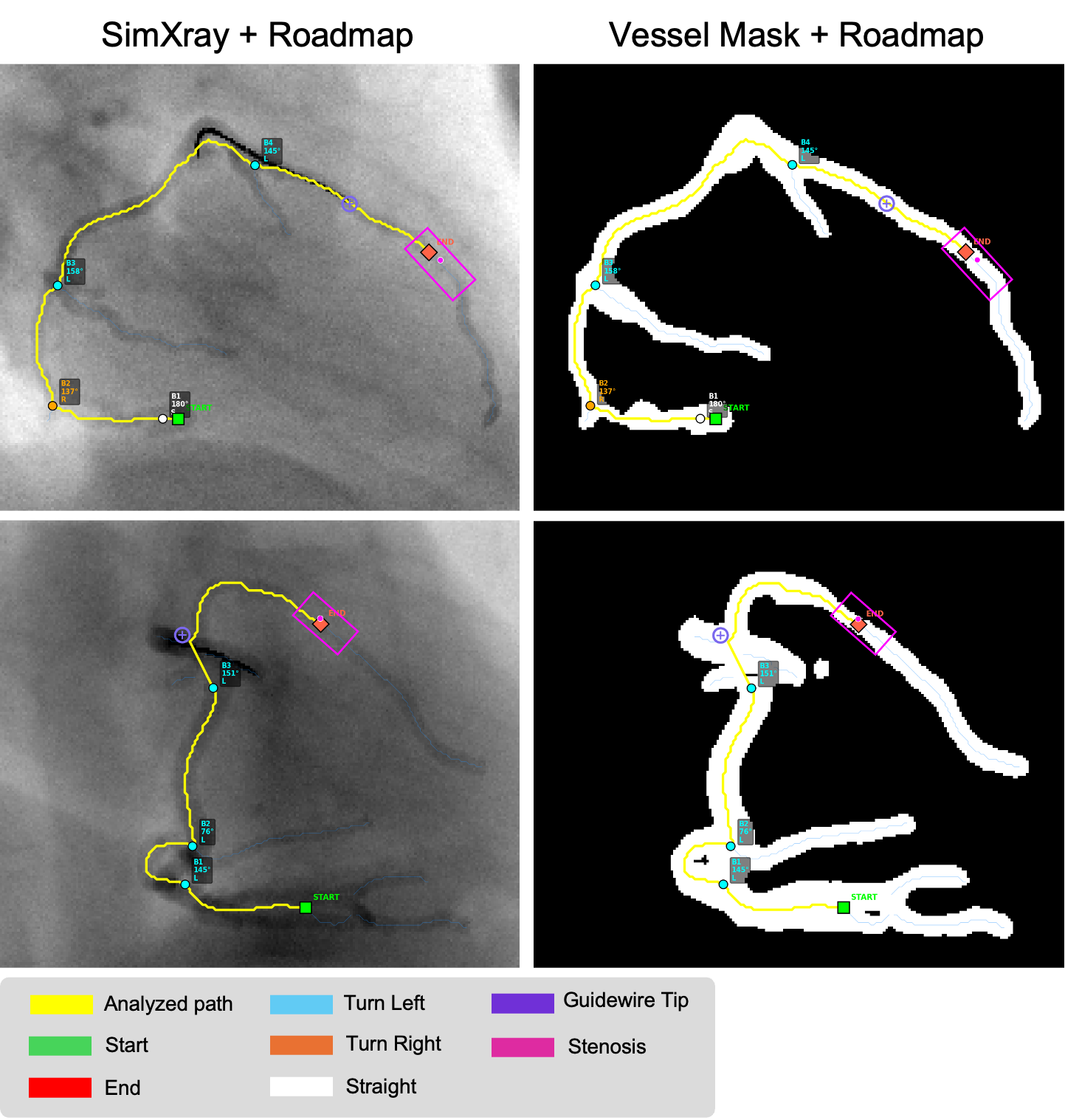}
    \caption{Real-time roadmap visualization for two representative cases. The root-to-stenosis path is overlaid on the simulated fluoroscopy image (left) and predicted vessel mask (right), together with the guidewire tip, stenosis, start and end points, and bifurcation-level navigation cues.}
    \label{fig:roadmap}
\end{figure}

Finally, Fig.~\ref{fig:roadmap} demonstrates the complete perception-to-roadmap workflow. The predicted vessel representation and stenosis location are used to determine a root-to-target centerline path, while the traversed bifurcations are annotated with branch angles and left/right/straight navigation cues. The roadmap is displayed both on the simulated fluoroscopy image and on the extracted vessel representation, providing an interpretable output that combines the learned perception results with the subsequent geometric analysis.

The complete real-time pipeline, including MCP-Net inference, vascular topology extraction, and roadmap generation, achieved a mean end-to-end latency of $26.90$ ms per frame, corresponding to $37.2$ FPS on the NVIDIA GB10 platform. This demonstrates that the complete perception-to-roadmap pipeline can operate at real-time rates.

\section{Discussion}
\label{sec:discussion}

The results show that MCP-Net can jointly recover vessel structure, guidewire-tip location, and stenosis information from a single simulated fluoroscopy frame. The high clDice score indicates good preservation of vessel connectivity, which is particularly important because missing or disconnected branches can directly affect the subsequent graph and navigation path. Guidewire-tip localization was robust, whereas stenosis localization exhibited a heavier error tail, indicating that precise lesion localization remains the more challenging perception task.

The proposed framework is particularly relevant to robotic-assisted interventions because it converts image-level predictions into an explicit vascular graph and target-oriented roadmap. Centerline-based representations have previously supported autonomous endovascular navigation ~\cite{li2024vascular}, while simulation has enabled guidewire-navigation strategies across varying vascular geometries ~\cite{scarponi2024zero,scarponi2024autonomous}. Our framework complements these approaches by deriving vessel geometry, device location, and the target directly from fluoroscopic perception. Since simulation provides known anatomy, imaging geometry, and automatically generated supervision, the same principle could also be extended to other image-guided robotic procedures, such as TAVI or bronchoscopy, using task-specific anatomical representations.

The main limitation is that training and evaluation are currently restricted to simulated fluoroscopy. Although simulation enables controlled variation and exact geometric supervision, a domain gap to clinical images remains. Simulation-to-real approaches such as XA-Sim2Real~\cite{zhang2024xa} provide a possible direction for addressing this limitation. Future work will therefore focus on phantom and clinical validation, simulation-to-real adaptation, and integration of the generated roadmap with closed-loop robotic guidewire control.

\section{Conclusion}
\label{sec:conclusion}

This work presented a simulation-driven framework for real-time coronary perception, vascular topology analysis, and roadmap generation from X-ray fluoroscopy. MCP-Net jointly estimates the coronary vessel structure, guidewire-tip location, and target stenosis, while subsequent geometric processing transforms these predictions into an interpretable vascular graph and root-to-target navigation roadmap. Guidewire-tip localization additionally provides information about the current device position relative to the vessel centerline, which can support guiding the guidewire along the generated roadmap toward the target lesion. The results demonstrate the feasibility of integrating learned fluoroscopic perception with structured vascular reasoning to provide navigation-relevant information for robotic-assisted PCI.

\bibliographystyle{IEEEtran}
\bibliography{reference}

\end{document}